\documentclass[letterpaper,10pt,conference]{ieeeconf}  

\IEEEoverridecommandlockouts                              

\usepackage{graphics} 
\usepackage{epsfig} 
\usepackage{mathptmx} 
\usepackage{times} 
\usepackage{amsmath} 
\usepackage{amssymb}  

\usepackage{multirow}
\usepackage{soul}
\usepackage{pifont} 
\usepackage{makecell}
\newcommand{\loss}{\mathcal{L}}
\usepackage{amsfonts}

\usepackage[utf8]{inputenc}
\usepackage[ruled,vlined]{algorithm2e}
\usepackage{color}
\usepackage{cite}
\usepackage{booktabs}

\newcommand{\R}{\mathbb{R}}

\newcommand{\vlnbert}{VLN\kern 0.13em $\circlearrowright$\kern 0.13em BERT}

\newcommand{\winmodel}{\textbf{W\kern 0.08em I\kern 0.08em N}}
\newcommand{\high}[1]{{\textbf{\color{blue}#1}}}
\newcommand{\secondhigh}[1]{{\textbf{\color{red}#1}}}

\usepackage{breqn}
\usepackage{algpseudocode}

\usepackage{color,soul}

\title{\LARGE \bf
What Is Near?: Room Locality Learning for Enhanced Robot Vision-Language-Navigation in Indoor Living Environments 
}

\author{Muraleekrishna Gopinathan$^{1}$, Jumana Abu-Khalaf$^{2}$, David Suter$^{3}$, Sidike Paheding$^{4}$, and Nathir A. Rawashdeh$^{5}$
\thanks{Muraleekrishna Gopinathan, Jumana Abu-Khalaf and David Suter are with the School of Science, Edith Cowan University, Joondalup, WA 6027, Australia (email: {\tt\small k.gopinathan@ecu.edu.au; j.abu-khalaf@ecu.edu.au; d.suter@ecu.edu.au}) \textit{corresponding author: Muraleekrishna Gopinathan}}%
\thanks{Sidike Paheding is with the Department of Computer Science, Fairfield University, 1073 N Benson Rd, Fairfield, CT 06824, USA (email: {\tt\small spaheding@fairfield.edu}}
\thanks{Nathir A. Rawashdeh is with the Department of Applied Computing, Michigan Technological University, 1400 Townsend Drive, Houghton, Michigan 49931-1295, USA (email: {\tt\small spahedin@mtu.edu; narawash@mtu.edu}}%
}
\begin{document}

\setlength{\textfloatsep}{6pt}

\maketitle
\thispagestyle{empty}
\pagestyle{empty}

\begin{abstract}

Humans use their knowledge of common house layouts obtained from previous experiences to predict nearby rooms while navigating in new environments. This greatly helps them navigate previously unseen environments and locate their target room. To provide layout prior knowledge to navigational agents based on common human living spaces, we propose WIN (\textit{W}hat \textit{I}s \textit{N}ear), a commonsense learning model for Vision Language Navigation (VLN) tasks. VLN requires an agent to traverse indoor environments based on descriptive navigational instructions. Unlike existing layout learning works, WIN predicts the local neighborhood map based on prior knowledge of living spaces and current observation, operating on an imagined global map of the entire environment. The model infers neighborhood regions based on visual cues of current observations, navigational history, and layout common sense. We show that local-global planning based on locality knowledge and predicting the indoor layout allows the agent to efficiently select the appropriate action. Specifically, we devised a cross-modal transformer that utilizes this locality prior for decision-making in addition to visual inputs and instructions. Experimental results show that locality learning using WIN provides better generalizability compared to classical VLN agents in unseen environments. Our model performs favorably on standard VLN metrics, with Success Rate 68\% and Success weighted by Path Length 63\% in unseen environments. 


\end{abstract}

\section{INTRODUCTION}
 Vision-Language Navigation (VLN) requires an agent to traverse through indoor environments based on descriptive navigational instructions. This task has garnered significant interest from both the computer vision and natural language processing (NLP) research communities, due to its practical applications in domestic settings. A VLN agent must learn to align visual inputs and instructions to execute a series of actions in order to reach a target location or object \cite{anderson2018vision}. This is different from classical goal navigation problems, as task success is measured by how well agent's trajectory conforms to the instruction. This task is inherently complex, as both natural language and visual understanding are challenging on their own. To perform this task, the agent must have a Natural Language (NL) model to \textit{understand} instructions, a vision model to extract visual features, and a model to learn the visual-language correspondence in order to determine the best action at each step. The agent must also keep track of its navigational history to estimate its progress. 
 
 \begin{figure}[t]
  \centering
  \includegraphics[width=\columnwidth]{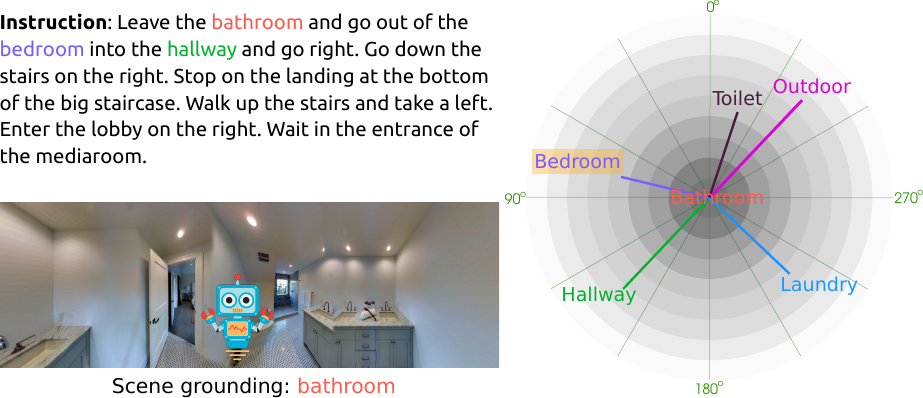}
  \caption{Using ego-centric layout knowledge for navigational reasoning. Agent grounds the token  \textit{bathroom} to current location, predicts neighboring scenes and decides the best action based on the rest of the instruction.}
  \label{fig:egocentric}
\end{figure}

 Existing VLN agents utilize SoTA vision and language models to encode visual cues and language instructions, and perform cross-modal encoding to learn their relationship. However, due to the variability in the appearance of home environments, these models perform poorly in previously unseen environments compared to seen ones. Previous works have attempted to improve performance through improving vision-language cross-modal learning \cite{Landi2019a,zhu2021multimodal}, training data augmentation \cite{li2022envedit, Liu2021REM, Tan2019EnvDrop}, and applying mixture of training methods \cite{Wu2021RL,Li2020URL}, but there remains a significant gap in navigational success between agents in seen and unseen environments.

These methods overlook the inherent patterns in human living spaces, such as the proximity of bathrooms to bedrooms. An agent can more effectively navigate an environment if it knows the relative locations of rooms, or the room-to-room relationships. To address this, we propose a new model that learns these layout patterns during training and predicts nearby room categories during validation. These predictions, along with the agent's understanding of the instructions, are used to select the next action.
 
Some recent work have used ego-centric maps for short-distance navigation by predicting a semantic map of visible regions \cite{Narasimhan2020Seeing} \cite{cartillier2020semantic}. However, these methods tend to have redundant information in their egocentric occupancy maps and encapsulate limited information about the surrounding area. This is evident from their limited navigational success even when ground truth maps are provided. Additionally, these methods have not been tested against long-horizon tasks, such as the Room-to-Room (R2R)\cite{anderson2018vision} and REVERIE \cite{qi2020reverie} tasks, which involve complex instructions and span multiple locations.

To address these limitations, we propose the \winmodel: '\textbf{W}hat \textbf{I}s \textbf{N}ear' approach for vision-language navigation. The WIN model is trained on neighborhood adjacency data extracted from large real-world house environments, such as Habitat 3D \cite{Savva2019HabitatAP}. Given a panoramic image of a room, the WIN agent predicts the categories (classes or \textit{types}) and relative locations of surrounding regions based on the visual cues from the current view (Fig. \ref{fig:egocentric}). The navigational agent then uses this local knowledge and language understanding to select the next action from the available directions in the view. We hypothesize that incorporating layout information will significantly reduce the number of probable action decisions obtained using only vision and language modalities. The compact representation of the local neighborhood used by the model also simplifies training and validation.

We evaluate the performance of our WIN model on the Room-to-Room (R2R) \cite{anderson2018vision} and REVERIE\cite{qi2020reverie} datasets. Experimental results show that learning layout prior using WIN improves the generalization ability of baseline agents in unseen environments. The model performs favorably on standard VLN metrics: Success Rate (SR)  (measures how well agent reaches target location) is $68\%$, and Success Weighted by Path Length (SPL)  (measures how well agent reaches target using shortest paths) is $63\%$ - on unseen environments.  We successfully demonstrate that layout prior knowledge can reduce the performance margin between seen and unseen environments.

Our contribution in this paper is as follows:
\begin{itemize}
    \item We propose a novel next-action-reasoning model for VLN based on \textit{locality} prior knowledge, conditioned on current view and navigational history (Sec. \ref{sec:winmodel}).
    \item We develop a mechanism to transform agent-centric locality knowledge to global map for trajectory prediction (Sec. \ref{par:targetencoder}). 
    \item We demonstrate that locality knowledge improves the performance of simple and classical VLN agents, without adding significant complexity (Sec. \ref{sec:results}).
\end{itemize}
\section{Related Work}
\subsection{Data Augmentation}
Recent studies in VLN have developed additional training data improving generalization. Early attempts synthesized additional training data; by back-translating instructions from trajectories \cite{fried2018speaker,wang2022less}, by mixing parts of existing paths \cite{Liu2021REM,Jain2020r4r}, and by generating images and instructions from web \cite{guhur2021airbert}. Generating visual variations in the environment were studied in \cite{Tan2019EnvDrop,li2022envedit} to improve the navigational performance in unseen environments. A recent study \cite{Chen2022HM3DAutoVLN}, automatically generated navigational graphs and instructions from un-annotated large scale Habitat-Matterport 3D (HM3D) \cite{ramakrishnan2021hm3d} dataset aiming to provide more training examples to VLN agents. This dataset has multiple viewpoints in the same room and focuses on navigability among them. In our work, we develop our \textit{Locality Knowledge} from un-annotated HM3D dataset but extract inter room-visibility and relative room locations. Unlike  \cite{Chen2022HM3DAutoVLN}, we select one viewpoint per room for room type inference and cleverly deduce visibility (occlusion) and locality (shares a wall) among rooms.


\subsection{Environment Map Learning}
 Learning to perform navigation in indoor spaces requires the agent to build and update a locality map representation of the environment. Generalization of navigational experiences to complex and unseen environments motivated recent works in end-to-end learning-based approaches \cite{Georgakis2022CM2, Zhu2017Targetdriven, Narasimhan2020Seeing}. Recently, a transformer-based topological planner \cite{Chen2021Topological} that uses graph neural network features to encapsulate the relationship between location connectivity and NL instructions was introduced. The limitation of this method is that the agent has to pre-explore the environment and build a map before attempting the VLN task.
 
 Semantic mapping for visual, vision-and-language, and point goal navigation tasks has been studied for continuous environments (VLN-CE). Existing approaches in the PointNav task \cite{Savva2019HabitatAP}, which is a short-horizon task, use Knowledge Graphs to represent room-object relationships in an environment during training and match the learned relationships to the test environments during inference \cite{yang2018visual}. For language-based navigational tasks, semantic map generation and environment layout learning are still unexplored. Our work is analogous to \cite{Georgakis2022CM2} in that they predict object locations and direction in an ego-centric map. Similar studies failed to generalize well in unseen environments as they map the visible regions only and not the occluded neighborhoods \cite{ramakrishnan2020occant}, have limited region classes \cite{Liang2021SSCNav} and suffer from room-object contextual-bias in the object-centric methods \cite{Liang2021SSCNav} (i.e. Tables can be in different rooms) . Another work \cite{Narasimhan2020Seeing} that uses similar indoor environment priors, which is applied in the PointNav problem  cannot be re-purposed for the VLN task because of its limited region categories and difference in task complexity. As the VLN task has a long-horizon trajectory, instruction, and intermediate goals, it is pertinent that the neighborhood map contains a larger number of room categories, conforms to the instruction, and accentuates the action decision at each navigational step. 
 
 In this work, we propose to learn room connectivity information (including both visible regions and regions that are occluded but adjacent to the current room) directly from the environment during training and apply this knowledge to improve action decisions. 

\section{Method}

In this section, we introduce the Vision-language Navigation problem, our methods and delineate on building the locality knowledge dataset.

\subsection{Problem Setup}
In VLN, an agent is placed in a discrete indoor environment in which each location represents a node of a predefined connectivity graph. Given a natural language instruction describing a trajectory from the start location to a target location, an agent at every time step perceives a panoramic RGB-D view of its surrounding and chooses the next viewpoint through an action direction from candidate viewpoints (navigable or unobstructed directions in a panoramic viewpoint). The agent executes a sequence of actions to complete the instruction until it decides to stop, ideally at the goal location, completing the episode.

\subsection{Overview of the WIN Model}
Learning room layout  patterns in navigational spaces can support robots to efficiently reach goal locations. 
During the training phase, the agent learns to extract connectivity information between rooms types (i.e. \textit{bathroom, bedroom, toilet}) and their relative orientations from \textit{Locality Knowledge} (Sec. \ref{sec:tkb}) w.r.t to the agent heading. This knowledge encodes different room-to-room connections (such as through doors, hallways or walls), room types, their relative locations, and distances to room centres. The agent learns visual-language correspondence along with topological relations that exist in the environment at each step of the navigational episode. Later, in the navigation phase, the agent uses the knowledge associated with each viewpoint to evaluate the action choices. Hence, the final action decision will be based on the agent's instruction-understanding, visual grounding of the current observation, and locality knowledge. The WIN model is detailed in Sec. \ref{sec:winmodel}.

We adopt a modular approach to provide the locality knowledge to the VLN agent. Specifically, we build a simple WIN model such that it can be added to existing VLN models to improve navigational success.

\subsection{Model Inputs}

\subsubsection{Language Encoding}
An instruction of $n$ words $X = \langle x_1, x_2,...,x_n\rangle$ is given to the agent at the start of a navigational episode. This instruction is tokenized and applied to a language encoder to obtain a language feature. 

\subsubsection{Vision Embedding}
A single trajectory consists of a sequence of $K$ panoramic views (steps) $V = \langle V_1, V_2,...,V_K \rangle$, each comprising of 36 single views in 3 camera elevations (up, horizon and down). At each time step $t$, we extract RGB-D visual feature $I_i$ and depth feature $D_t$ as visual context. Also, we add the relative heading $\theta$ and elevation $\phi$ angles of each view with respect to the agent's current orientation to retain view directions $R_t = [\cos{\theta}, \sin{\theta}, \cos{\phi}, \sin{\phi}]$.

\subsection{Locality Knowledge}
\label{sec:tkb}

\begin{figure}[tbp!]
\centering
  \includegraphics[trim={0 0.1cm 0 5px},clip,width=0.90\columnwidth]{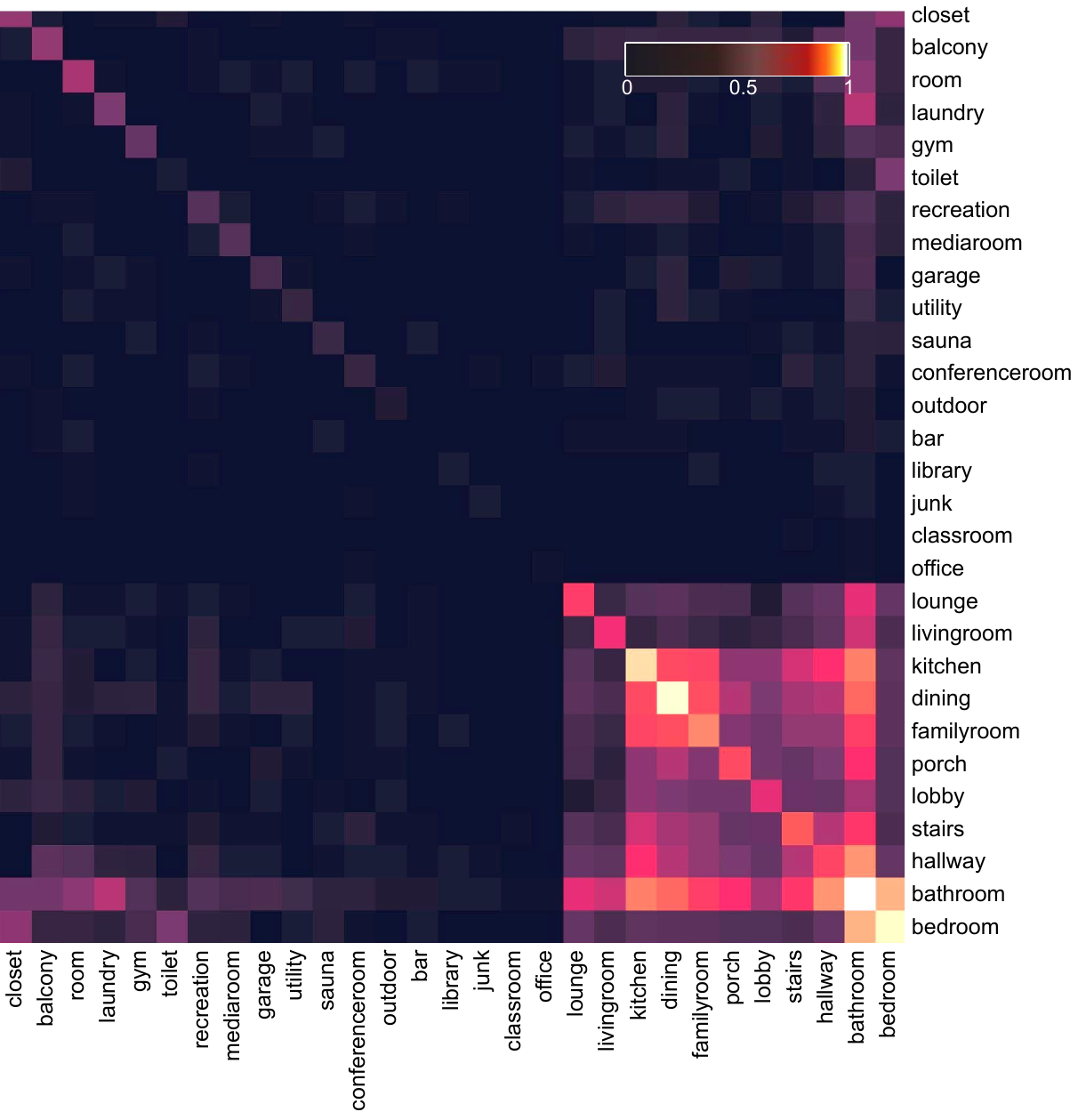}
  \caption{Room adjacency matrix of Habitat-Matterport 3D (HM3D) dataset. Each cell represents connectivity (share a wall), navigability (direct access) or visibility (line-of-sight) between room types. Brighter colours show large co-occurrence of room types in their neighbourhood.}
  \label{fig:topomapping}
  
\end{figure}
The core of locality prediction includes the representation of the local neighbourhood, predicting the ego-centric locality map based on common patterns and improving the action probability using the predicted map. 

To predict room layouts, we trained a our \textit{Locality Predictor} using HM3D dataset which has 900 houses with room panoramas, camera poses, and floor plans. The HM3D dataset has realistic 3D indoor scenes but lacks the navigability graphs, or region labels or room boundaries. Hence, to build the \textit{Locality Knowledge} base, we need 1) one summary viewpoint per room, 2) its type and 3) geodesic distance and orientation between neighboring rooms. The summary viewpoints are collected by sampling equidistant viewpoints from navigable regions of the scene and eliminating points based on following conditions. To obtain single viewpoint per room;  (1) candidate viewpoints should be at least 2 meters apart from each other (2) the panoramic images from viewpoints cannot have significant matching ORB descriptors \cite{Rublee2011ORB} to eliminate candidates from same room and, (3) farther viewpoints with matching descriptors are navigable from each other, and (4) close viewpoints with fewer to no matches are considered neighboring but occluded from each other. A ResNet \cite{He2015DeepRL} model trained on the MP3D  \cite{chang2017matterport3d} dataset was used to obtain the room types. The best thresholds for ORB descriptor matching are selected for the maximum coverage of the scene. The distances are based on average room sizes and room-to-room distances in the HM3D dataset. The room adjacency matrix (Fig. \ref{fig:topomapping}) shows common room neighborhoods, connectivity and adjacency. 

The resulting \textit{Locality Map} $M^{GT}$ is a metric-semantic map of the region surrounding the agent. That is, the area around the agent is divided into a fixed-size grid $g\times g$ with a side $s$ to represent the region of interest. Each cell encodes the location, orientation and type of the rooms with respect to the heading of the agent.

\begin{figure}[tbp]
\centering
    \includegraphics[width=.95\columnwidth]{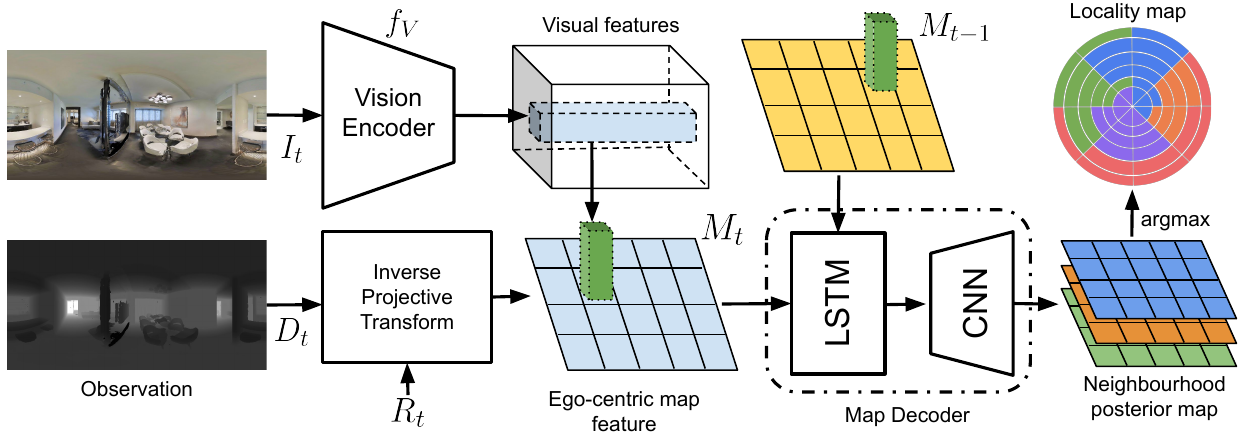}
    \caption{Locality predictor produce an ego-centric locality map $M_t$ based on RGB-D input and past action $a_t-1$. The visual feature $f_V(I_t)$ is projected to map feature space using $D_t$ and orientation $R_t$. The locality Map decoder uses LSTM to integrate robot motion and predicts new locality maps based on the current $M_t$ and previous $M_{t-1}$ ego-centric maps.}
    \label{fig:locality_predictor}
\end{figure}

\subsection{Locality Predictor}
The locality predictor (Fig. \ref{fig:locality_predictor}) uses horizon visual features and predicts room class (type) of the grid area surrounding the agent. This module uses the agent's panoramic observation to produce a probability distribution of region categories for each cell in the locality map. The predictor contains two functions namely (1) egocentric mapper and (2) neighborhood predictor.
The former is an affine transformation and inverse projective mapping of the semantic visual features to the ground plane to obtain the current map using the camera parameters, and visual inputs. To suppress feature collision during ground projection, we follow MapNet \cite{Henriques2018MapNet} and take the maximum of values height-wise. Evidently, this map only includes the regions that are visible to the agent. The neighborhood predictor network is trained to extend this projected map to invisible regions using supervision. The \textit{Map Decoder} concatenates current $M_t$ and previous $M_{t-1}$ maps using a trainable network with parameter $W_M$ to obtain map feature $m_t$,

\begin{equation}
    m_t = [M_{t-1}; M_t] W_M 
\end{equation}
and applies $m_t$ to LSTM to track the map evolution due to previous agent action in the hidden feature $h_t$:
\begin{equation}
        h_t = LSTM([m_t; \hat{a}_{t-1}], h_{t-1})
\end{equation}

where $\hat{a}$ is the action embedding.
The updated map is the probability distribution of room types for each direction,
\begin{equation}
        p_{M,t} = softmax(f_v W_M h_t)
\end{equation}    


\subsection{Target Encoder}
\label{par:targetencoder}

\begin{figure}[tbp]
\centering
    \includegraphics[width=0.90\columnwidth]{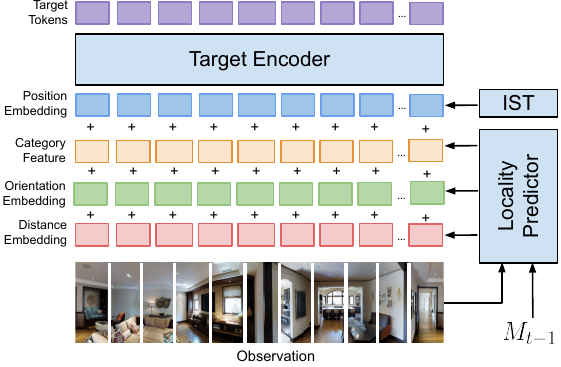}
    \vspace{-0.5em}
    \caption{Neighbourhood encoding utilizes view embedding, orientation embedding, room category feature and position}
    \label{fig:neighbourhood_encodding}
\end{figure}
In order to utilise the locality predictions for action
selection, we transform the locality map to global grid. For this we obtain position token of each target location in the global grid from Imaginary Scene Tokenization (IST) mechanism in \cite{Zhao2022TDSTP} to provide the global map. Each target token represents prediction of the scene layout in its cell. To this we add room type feature, orientation and distance of each view of the panorama to obtain target tokens $c_t$ (Fig. \ref{fig:neighbourhood_encodding}).  These target tokens are applied to the structured transformer to generate updated target tokens which is recursively applied to the transformer in the next time step.

\section{What Is Near (\winmodel)~Model}
\label{sec:winmodel}

 \begin{figure*}[tbp]
  \centering
  \includegraphics[width=0.98\linewidth]{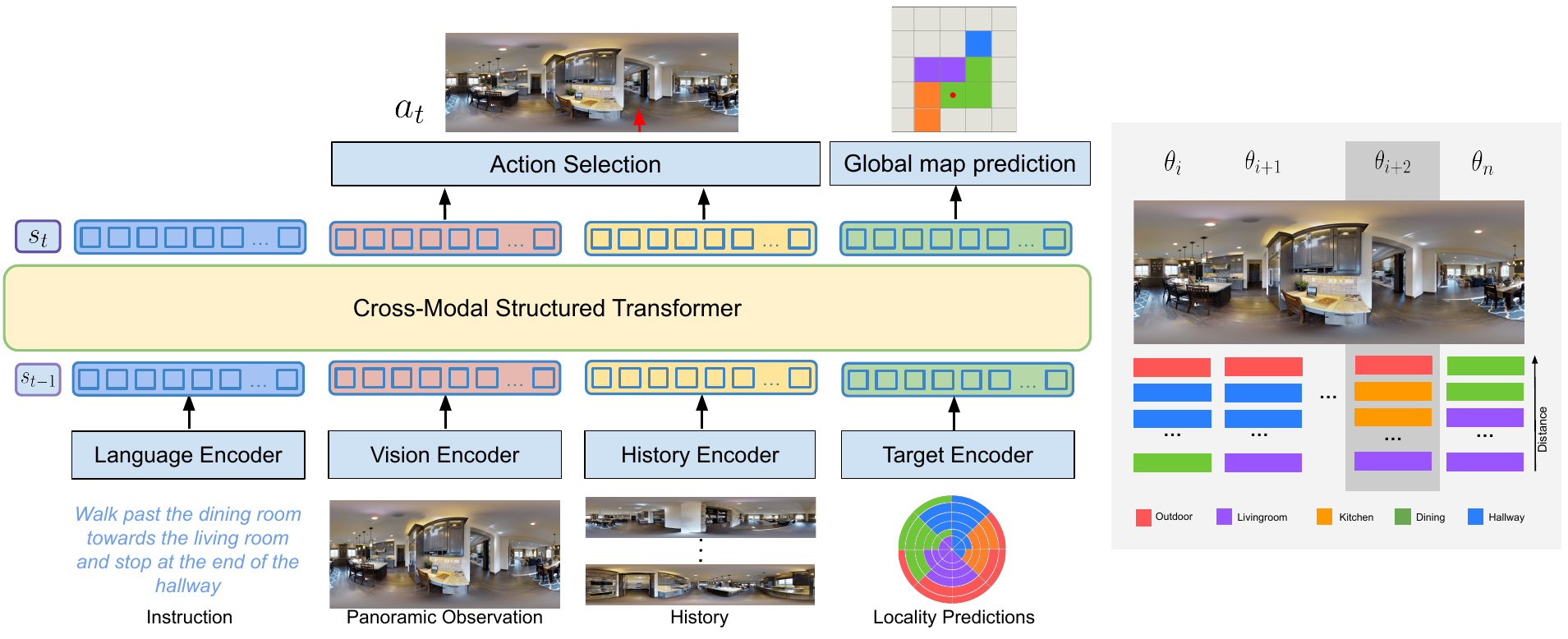}
  \caption{The proposed What is Near (WIN) model (left) and representation of egocentric locality map (right). The cross-modal structured transformer accepts language, vision, history and target tokens to predict action probabilities and semantic map of the global space. A hidden state $s$ vector encodes the state and history of navigation episodes while location encoder combines global target tokens with locality embedding. This locality map (ego-centric) to global map transformation provides agent the long-term planning capability}
  \label{fig:architecture}
\end{figure*}

The What Is Near (WIN) model (Fig. \ref{fig:architecture}) is composed of a cross-modal transformer that is adapted from \cite{chen2021hamt}. At each time step $t$, the model takes in inputs from previous state $s_{t-1}$, language encoding $X$, vision embedding $V$, history tokens $h_{t-1}$, and a neighborhood encoding $N_t$ which captures the agent's knowledge of the current scene. The language encoding is time-independent, while the state and history tokens come from previous time steps, and the visual and neighborhood tokens are obtained for the current scene. The WIN model processes the entire language instruction and performs self-attention with the help of a BERT language encoder, generating an initial state token $s_0$ and language embedding. A vision encoder $f_V$ is used to encode the panoramic scene and produce visual features. The cross-modal transformer then performs cross-attention on the language, observation, history, and target tokens to learn their correspondence, and the \texttt{[CLS]} embedding of the transformer is used to predict the action.

The WIN model leverages the panoramic image visual features to predict the local neighborhood, locations, and action probabilities by considering prior knowledge and visual cues from the scene. The action predictor operates based on the visual-language-locality correspondence learned by the cross-modal transformer and produces a probability distribution over each candidate direction.

We extend the Structure Transformer Predictor (STP) mechanism from \cite{Zhao2022TDSTP} to model the overall space of the environment.  We adapt this model for our task as a region category prediction task. The history token of the STP is composed of,

\begin{equation}
    H_t = f_V(I_t) + f_R(R_T) + f_T(t) + f_P(l_t)
    \label{eq:history-stp}
\end{equation}

\noindent where $f_R, f_T, f_P$ are trainable encoders for motion direction, navigation step, and agent global position respectively. $f_V$ is the encoder for panoramic view $I_t$. To provide the locality map to the transformer, we need to extent the local map to the global coordinate system. For this we transform the locality prediction from agent-centric coordinate system to global map using rigid-body transformation. This transformation also maps locality map to the global grid. We reuse IST of the STP model and generate target tokens $c_i$ for each global grid cell such that,

\begin{equation}
    c_i = f_P(l_t) * x_0 * f_M(I_t, D_t, M_{t-1})
\end{equation}

where $f_P(I_t)$ is the same position encoder as in (\ref{eq:history-stp}), $x_0$ is the instruction embedding, and $f_M$ is the local-global transformation. 
 \textit{Locality predictor} $f_M$ has recurrent memory model based on LSTM trained on locality information. For details of STP, we encourage readers to refer to \cite{Zhao2022TDSTP}.

Finally, to produce action probabilities based on the learned locality knowledge, we formulate action prediction as a classification problem. An MLP is applied to the vision-language representation to predict an action probability distribution over navigable viewpoints as in \eqref{eq:actionpredicition}. 
\begin{equation}
\label{eq:actionpredicition}
    p_t(I^p_i) = \frac{\exp{f_A(I^p_i \odot ^{VL}c_t)}}{\sum_j{\exp{f_A(I^p_j \odot ^{VL}c_t)}}}
\end{equation}

where $\odot$ represents element-wise multiplication and $^{VL}c_t$ is the vision-language fused representation. As in existing works that use pretrained vision-language models, we use the embedded \texttt{[CLS]} token that is a fused representation of vision-language modalities \cite{Hong2021recurrent} as the state representation. In all, the complete model is \eqref{eq:applytopo},

\begin{equation}
    \label{eq:applytopo}
    s_t, p^a_t = \textsc{WIN}(h_{t-1}, X, I^c_t, P_t, c_t)
\end{equation}

\noindent where $s_t$ is the state vector and $p^a_t$ is the action probability.

\subsection{Training}
\label{sec:training}
In this section, we describe the training procedure for the WIN model. Our model is trained in 2 parts: Locality Predictor module training and end-to-end training for VLN. 
\subsubsection{Locality Predictor} The Locality prediction $f_T$ module is trained using the Locality Knowledge (see Sec. \ref{sec:tkb}. The model is trained by providing the panoramic observation at different agent orientations for each scene and comparing the prediction with the ground truth. The objective is to minimize \eqref{eq:topoloss},


\begin{equation}
\label{eq:topoloss}
\loss_{locality} = \sum_{k\in\R} CrossEntropyLoss(M^{pred}_{t,k}, M^{GT}_{t,k})
\end{equation}

\subsubsection{Action Prediction}
We adopt a combination of Reinforcement Learning (RL) and Imitation Learning (IL) for training our agent. Imitation Learning is applied to train the agent while providing the ground truth action or \textit{teacher action} at each time step, and minimizing the cross-entropy loss defined by \eqref{eq:topoloss}. For RL, we use Advantage Actor Critic (A2C) \cite{mnih2016asynchronous} to learn actions that maximize rewards from reducing the distance to the goal location at each time step and arriving within 3m of the target, at the end of a navigational episode.

During navigational training, the target encoding $N_t$ from the frozen \textit{Locality Predictor} is used as an input to the cross-modal transformer along with history $H_t$ and vision-language inputs.  Following \cite{Zhao2022TDSTP} we include \textit{history teacher loss} to accommodate the change in action space with visited locations. The final loss aims to minimize the negative likelihood of the target view $I_{*,t}:\loss_A =-\log p_t(I^p_t) $ and the \textit{history teacher loss}. Formally we minimize for all steps $T$,

\begin{equation}
 \loss_{IL} = - \sum_{t=1}^T \pi \log (a_G^t; \Theta) - \sum_{t=1}^T \log p_t(I^p_t)
\end{equation}

\noindent where $a_G$ is the global action towards the goal, $\pi$ is the navigational policy parameterized by $\Theta$. Another MLP is used to decode the global target from the semantic target tokens for the global map prediction.

 The agent samples action $a^*_t$ from action probability $p^*_{a_t}$ from the \winmodel~model. We found that a combination of IL and RL balances exploration-exploitation strategies effectively: defined as,

\begin{equation}
    \loss_{RL+IL} = - \sum_{t=1}^T a^*_t \log(p^*_{a_t}) A_t + \lambda_{IL} \loss_{IL}  
\end{equation}

\noindent where $\lambda_{IL}$ is the IL training coefficient and $A_t$ is the advantage calculated by the A2C algorithm \cite{mnih2016asynchronous}.

\begin{table*}[htp!]
  \centering
  \caption{Comparison of agent performance on R2R in single-run setting. \high{Blue} and \secondhigh{Red} denote best and second best respectively. $\spadesuit$ works that use augmented datasets for training. $\clubsuit$ denotes pre-trained agents. }
  \begin{tabular}{l|cccc|cccc|cccc}
    \toprule
    \multicolumn{1}{c|}{\multirow{2}{*}{Methods}} & \multicolumn{4}{c|}{ValSeen} & \multicolumn{4}{c|}{ValUnseen} & \multicolumn{4}{c}{TestUnseen} \\
     & \multicolumn{1}{c}{TL} & \multicolumn{1}{c}{NE$\downarrow$} & \multicolumn{1}{c}{SR$\uparrow$} & \multicolumn{1}{c|}{SPL$\uparrow$} & \multicolumn{1}{c}{TL} & \multicolumn{1}{c}{NE$\downarrow$} & \multicolumn{1}{c}{SR$\uparrow$} & \multicolumn{1}{c|}{SPL$\uparrow$} & \multicolumn{1}{c}{TL} & \multicolumn{1}{c}{NE$\downarrow$} & \multicolumn{1}{c}{SR$\uparrow$} & \multicolumn{1}{c}{SPL$\uparrow$} \\
    \hline \hline
    Human      & -     & -    & -    & -    & -     & -    & -    & -    & 11.85 & 1.61 & 86 & 76 \\
    Seq2Seq-SF \cite{anderson2018vision} & 11.33 & 6.01 & 39 & - & 8.39  & 7.81 & 22 & -    & 8.13  & 7.85 & 20 & 18 \\
    PREVALENT$^\spadesuit$~\cite{hao2020towards} & 10.32 & 3.67 & 69 & 65 & 10.19 & 4.71 & 58 & 53 & 10.51 & 5.30 & 54 & 51 \\
    \vlnbert~$^\clubsuit$ \cite{Hong2021recurrent} & 11.13 & 2.90 & 72 & 68 & 12.01 & 3.93 & 63 & 57 & 12.35 & 4.09 & 63 & 57 \\
    AirBERT$^{\spadesuit\clubsuit}$~\cite{guhur2021airbert} & 11.09 & 2.68 & 75 & 70 & 11.78 & 4.01 & 62 & 56 & 12.41 & 4.13 & 62 & 57 \\
    
    HAMT~\cite{chen2021hamt} & 11.15 & 2.51 & 76 & 72 & 11.46 & \high{2.29} & 66 & 61 & 12.27 & 3.93 & 65 & 60 \\

    TD-STP$^\spadesuit$ \cite{Zhao2022TDSTP} & 12.74 & \secondhigh{2.34} & \secondhigh{77} & \secondhigh{73} & 14.71 & 3.22 & \secondhigh{70} & \secondhigh{63} & - & {3.73} & 67 & 61 \\
    EnvEdit$^\spadesuit$ \cite{li2022envedit} & 11.18 & 2.32 & \secondhigh{77} & \high{74} & 11.13 & 3.24 & 69 & \high{64} & 11.90 & \high{3.59} & \secondhigh{68} & \high{64} \\
    
    DUET~\cite{chen2022duet} & 12.32 & 2.28 & \high{79} & \secondhigh{73} & 13.94 & 3.31 & \high{72} & 60 & 14.73 & 3.65 & \high{69} & 59 \\ 
    \hline 
    \rule{0pt}{2.5ex}Ours (WIN+\vlnbert) & 11.24 & 2.63 & 76 & 71 & 11.86 & 3.11 & 65 & 60 & 11.93 & 3.93 & 64 & 59 \\
     Ours (WIN+STP) & 12.53 & \high{2.25} & \high{79} & \high{74} & 13.14 & \secondhigh{3.09} & \high{72} & \high{64} & 13.02 & \secondhigh{3.61} & \secondhigh{68} & \secondhigh{63} \\
    \bottomrule
  \end{tabular}

\label{tab:table_r2r_result}
\end{table*}

\begin{table*}[htp!]
\centering
\caption{Comparison with state-of-the-art methods on the REVERIE dataset.}

\begin{tabular}{l|ccccc|ccccccc}
\toprule
\multicolumn{1}{c|}{\multirow{3}{*}{Methods}}  & \multicolumn{5}{c|}{ValUnseen} & \multicolumn{5}{c}{TestUnseen} \\
& \multicolumn{3}{c}{Navigation} & \multicolumn{2}{c|}{Grounding} & \multicolumn{3}{c}{Navigation} & \multicolumn{2}{c}{Grounding} \\
 \cmidrule(lr){2-4}  \cmidrule(lr){5-6}  \cmidrule(lr){7-9} \cmidrule(lr){10-11}
 & \multicolumn{1}{c}{SR$\uparrow$} & \multicolumn{1}{c}{SPL$\uparrow$} &  \multicolumn{1}{c}{OSR$\uparrow$}  & \multicolumn{1}{c}{RGS$\uparrow$} & \multicolumn{1}{c|}{RGSPL$\uparrow$} & \multicolumn{1}{c}{SR$\uparrow$} & \multicolumn{1}{c}{SPL$\uparrow$} & \multicolumn{1}{c}{OSR$\uparrow$} & \multicolumn{1}{c}{RGS$\uparrow$} & \multicolumn{1}{c}{RGSPL$\uparrow$} \\ \hline\hline
Seq2Seq~\cite{anderson2018vision} & 4.20 & 2.84 & 8.07 & 2.16 & 1.63 & 3.99 & 3.09 & 6.88 & 2.00 & 1.58 \\
FAST-MATTN~\cite{qi2020reverie} & 14.40 & 7.19 & 28.20 & 7.84 & 4.67 & 19.88 & 11.6 & 30.63 & 11.28 & 6.08 \\
\vlnbert~\cite{Hong2021recurrent} & 30.67 & 24.90 & 35.20 & 18.77 & 15.27 & 29.61 & 23.99 & 32.91 & 16.50 & 13.51 \\
HAMT~\cite{chen2021hamt}&  32.95 & 30.20 & 36.84 & 18.92 & 17.28 & 30.40 & 26.67 & 33.41 & 14.88 & 13.08 \\ 
TD-STP\cite{Zhao2022TDSTP} &  34.88 & 27.32 & 39.48 & \secondhigh{21.16} & \secondhigh{16.56} & 35.89 & 27.51 & 40.26 & \secondhigh{19.88} & \secondhigh{15.40} \\ 
DUET \cite{chen2022duet} &  \high{46.98} & \high{33.73} & \high{51.07} & \high{32.15} & \high{23.03} & \high{52.51} & \high{36.06} & \high{56.91} & \high{31.88} & \high{22.06} \\ \hline
\rule{0pt}{2.5ex}WIN-STP (OURS) &  \secondhigh{37.93} & \secondhigh{30.64} & \secondhigh{41.13} & 20.17 & 15.51 & \secondhigh{42.19} & \secondhigh{31.06} & \secondhigh{47.12} & 18.16 & 14.83 \\

\bottomrule
\end{tabular}

\label{tab:reverie_results}
\end{table*}

\section{Experiments}

In this section, we elaborate on our research questions, experiments and our baseline agents. From our experiments we aim to understand the following,
\paragraph{How does locality knowledge affect VLN agent performance?}
Neighborhood knowledge learned by the agent must include both the metric and semantic layout of the locality. On one end, an agent may use information such as, if a direction leads outdoors or indoors or at the other end utilizes the complete room type information. To measure the expressiveness of locality map, we test the agent on different map types including one with random map prediction and one with ground-truth map provided.

\paragraph{How does the performance of our WIN model compare to existing work in VLN?}
As the existing works ignore regions beyond the view of the agent, the current SoTA agents can benefit from locality knowledge for the next action prediction. We compare our model performance against SoTA VLN agents. 

\subsection{Baseline Agents}
We select two robust but computationally simple agents as our baseline to show how locality knowledge can affect their environment awareness and eventually, their navigational success. 

\subsubsection{\vlnbert} For the first baseline, we use a simple Recurrent VLN-BERT (\vlnbert) \cite{Hong2021recurrent} agent with basic history representation. \vlnbert~uses the \texttt{[CLS]} token of transformer in recurrent fashion as navigational history. Our model incorporates locality knowledge in this baseline by multiplying the global  map prediction with the action probabilities. 

\subsubsection{TD-STP} TD-STP \cite{Zhao2022TDSTP} is a method for enabling action reasoning by providing an encoded global grid positions to pretrained cross-modal encoder and predicting a \textit{target} cell based on the given instruction. For this, TD-STP imagines a discrete global grid over the entire floor area initially, and update the target location at each time step. We extend the  Structured Transformer Planner (STP) mechanism in our the WIN model for global grid semantic mapping.

\subsection{Setup}
\subsubsection{Dataset}
We evaluate WIN using validation splits of the Room-to-Room (R2R) \cite{anderson2018vision} and REVERIE \cite{qi2020reverie} datasets. R2R dataset consists of 7k trajectories from 90 houses split into \textit{Train} (61 houses), \textit{ValSeen} (houses from train seen), \textit{ValUnseen} (11 houses not included in train seen split) and \textit{TestUnseen} (18 houses not part of other splits). Each trajectory has 3 fine-grained English instructions. The test unseen split trajectories are submitted to an online system for evaluation\footnote{R2R leader board: https://evalai.cloudcv.org/web/challenges/challenge-page/97/overview}. The online server reports all metrics used for our evaluation. REVERIE dataset contains high-level instructions and uses the same split for training and evaluation\footnote{REVERIE leader board: https://eval.ai/web/challenges/challenge-page/606/leaderboard/1683}.

\subsubsection{Evaluation Metrics}
We use the standard metrics for evaluating the agent’s performance on the R2R and REVERIE datasets.  In R2R, the standard metrics such as Trajectory Length (TL), Navigation Error (NE), Oracle Success Rate (OSR), Success Rate (SR) and Success Rate weighted by Path Length (SPL) are reported \cite{anderson2018vision,anderson2018evaluation}. 

In addition to these, REVERIE \cite{qi2020reverie} also evaluates Remote Grounding Success (RGS) to measure the success rate of locating the remote object and Remote Grounding Success weighted by Path Length (RGSPL) which rewards shorter path lengths.

\subsubsection{Implementation details}
The model is built on PyTorch and experiments are performed on an NVIDIA A6000 GPU. Our model is trained for 100k iterations with early stopping applied at the highest SPL to prevent over-fitting.  The final results are reported for grid size $g$ 10 and cell size $s$ 0.5m. The batch size is set to 8 and the learning rate is $1e-5$. The dropout is set to $0.5$ and AdamW optimizer is used for training. We develop two baselines using publicly available source codes and hyper-parameters are set as per the original models.


\subsection{Results}
\label{sec:results}

\subsubsection{Results on the R2R dataset}

Our WIN+STP agent improved performance over the TD-STP baseline by a large margin (Table \ref{tab:table_r2r_result}). The TD-STP baseline uses visual features and instructions for predicting the global action space which is essentially an occupancy tracking.  Instead, the WIN model predicts the room layout which is a useful for local action selection  and reducing the overall path length. This shows WIN has comparatively better local action selection  due to the  layout understanding. The overall reduction in the navigation error (3.73m $\rightarrow$ 3.61m) compared to the baselines suggests that the agent is being directed to takes better actions based on the locality knowledge.

\subsubsection{Results on the REVERIE dataset}
WIN also shows better SR in \textit{TestUnseen} split (Table \ref{tab:reverie_results}) compared to the baselines. The agent could utilize descriptive instructions and select correct navigational actions at each step leveraging the locality knowledge. This improved instruction and layout understanding, lead to a higher navigational success rate (35.89\%$\rightarrow$ 42.19\%) and SPL (27.51\%$\rightarrow$31.06\%).

\section{Discussion}

We compare our results with larger and more complex SoTA models to show that our relatively simpler model performs competently using locality knowledge. Our WIN model has a better SPL ($63\%$) than DUET \cite{chen2022duet} (SPL: $59\%$) and same SR as EnvEdit \cite{li2022envedit}($68\%$) on the R2R \textit{TestUnseen} split. Both these agents are trained using augmented datasets; EnvEdit is trained on changed visual appearances of MP3D scenes and DUET is trained on multiple auxiliary tasks to learn local and global topology encoding. Compared to these methods which are computationally complex, our model training is simpler as we make use of simple methods to extract neighborhood knowledge. This computational advantage makes our model's performance gain more significant. WIN performs better than the baseline model on REVERIE task on navigational metrics. The object grounding scores, RGS and RGSPL, are not improved because WIN model only considers the room-to-room relations and not object-to-room relations. Overall, the locality knowledge in our WIN method is advantageous for navigational agents. 

\subsection{Effect of varying mapping area}
Here we compare the SR of our WIN agent with various locality map resolutions (Table \ref{tab:r2r_gridcompare}). We see lower SR for extreme grid sizes and highest SR for 5x5 grid. This can be explained by the average room sizes ($3.16 m^2$) in the MP3D dataset \cite{ramakrishnan2021hm3d}. The average distance between viewpoints in the R2R dataset is 2.25$\pm$0.57m with one or more viewpoints of them occupying the same room. Hence the largest grid size that WIN can predict well is about 2 average sized rooms. As the map size increases beyond two rooms, the prediction accuracy drops and the agent may be misguided.

\begin{table}[tp!]
    \centering
    \caption{Success rates on varying grid sizes for 0.5m x 0.5m cell size}
    \begin{tabular}{c|c|c}
    \hline
         $g \times g$ &  ValSeen SR & ValUnseen  SR \\
         \hline
         $3 \times 3$ &  64.23 & 58.47 \\
         $5 \times 5$ & 79.76 & 72.29 \\
         $7 \times 7$ & 73.96 & 61.38\\
         $9 \times 9$ & 59.14 & 51.33\\
         \hline
    \end{tabular}
    \label{tab:r2r_gridcompare}
\end{table}

\subsection{Impact of locality knowledge}

\begin{table}[tp!]
    \centering
    \caption{Success rates on various map types for 5x5 grid}
    \begin{tabular}{c|c|c|c}
    \hline
        Method & Map &  ValUnseen SR & ValUnseen SPL  \\
         \hline
         \#1 & Rand. type+Rand. dir. &  31.40 & 28.63 \\
         \#2 & Rand. dir.+GT type &  49.21 & 35.18 \\
         \#3 & Pred. type+Pred. dir. & 72.29 & 64.37 \\
         \#4 & LK GT & 78.22 & 67.16 \\
         \hline
    \end{tabular}
    \label{tab:r2r_gtcompare}
\end{table}
To measure the lower and upper bounds of WIN map prediction, we compare the SR with different types of maps provided to the agent in Table \ref{tab:r2r_gtcompare}.

We test four types of maps with grid size 5x5: Method \textit{\#1} for random room types (Rand. type) and directions (Rand. dir.), \textit{\#2} with random room location with room types from the locality knowledge ground-truth (GT type), \textit{\#3} with map prediction from the Map Predictor module (Pred. type) and \textit{\#4} with full ground-truth locality knowledge (LK GT). Method \textit{\#1} represents the lower bound performance of the WIN model resulting in lowest SR because the direction and room types in the map do not correlate with the environment. Also, a large SR-SPL 
margin of \textit{\#2} shows that the agent can still deduce the neighborhood but chooses wrong directions and takes longer trajectories.
In the upper-bound scenario, \textit{\#4}, the agent utilizes the actual room type and direction and obtains the highest SR and SPL.

\subsection{Limitations}
We observe that the performance of our WIN model degrades on trajectories with uncommon room types \textit{i.e.} uncorrelated classes shown by dark colours (Fig. \ref{fig:topomapping}). In certain failure cases the agent loses confidence in vision-language based action predictions and gets stuck in some location when the locality suggests diametrically opposite actions. This could be tackled by using locality knowledge from large house plan datasets with various room-to-room relationship examples. In future work, we plan to explore large-scale locality learning from real house plans.
\section{Conclusion}
We present a novel approach for VLN based on using room locality knowledge to predict neighboring rooms in the indoor environments. Our modular WIN model demonstrates a significant performance gain in unseen environments compared to the SoTA baselines. In this study, we encode layout patterns commonly found in indoor environments using a locality prediction model and use this knowledge to assist Vision-language navigation agents in making action decisions at each time step. Navigational results on the R2R and REVERIE tasks show that the WIN method outperforms both baseline methods while reducing the success rate margin between seen and unseen environments. A potential extension of this work is to learn general topological relationships from large-scale house plan datasets such as CubiCasa5k\cite{Kalervo2019cubicasa} and ZInD \cite{Cruz2021ZInD}.

\bibliographystyle{IEEEtran}
\bibliography{IEEEabrv,bibliography}

\end{document}